\documentclass[10pt,twocolumn,letterpaper]{article}

\usepackage{cvpr}      
\usepackage{times}
\usepackage{graphicx}
\usepackage{overpic}
\usepackage{amsmath}
\usepackage{amssymb}
\usepackage{svg}
\usepackage{amsmath}
\usepackage{algorithm}
\usepackage{listings}

\usepackage{etoolbox}
\usepackage{multirow}

\usepackage{makecell}

\definecolor{cvprblue}{rgb}{0.21,0.49,0.74}
\usepackage[pagebackref,breaklinks,colorlinks,citecolor=cvprblue]{hyperref}

\usepackage{overpic}
\usepackage{booktabs}
\usepackage{multirow}
\usepackage{pifont}
\usepackage{enumitem}
\usepackage{bm}
\usepackage{diagbox}

\usepackage{algorithm}
\usepackage{algorithmic}
\usepackage[accsupp]{axessibility}

\usepackage{tcolorbox}
\newtcolorbox[auto counter, number within=section]{namedbox}[2][]{
    colback=white,
    colframe=black,
    fonttitle=\bfseries,
    title=Box~\thetcbcounter: #2,
    #1
}

\definecolor{cvprblue}{rgb}{0.21,0.49,0.74}

\def\paperID{} 
\def\confName{CVPR}
\def\confYear{2026}

\title{FREE-Switch: Frequency-based Dynamic
LoRA Switch for Style Transfer}

\author{Shenghe Zheng, Minyu Zhang, Tianhao Liu, Hongzhi Wang\\
Harbin Institute of Technology\\
{\tt\small shenghez.zheng@gmail.com}, {\tt\small wangzh@hit.edu.cn}
}

\begin{document}

\twocolumn[{%
\renewcommand\twocolumn[1][]{#1}%
\maketitle
\begin{center}
   \captionsetup{type=figure}
    \vspace{-1.0cm}    
\includegraphics[width=1.0\linewidth]{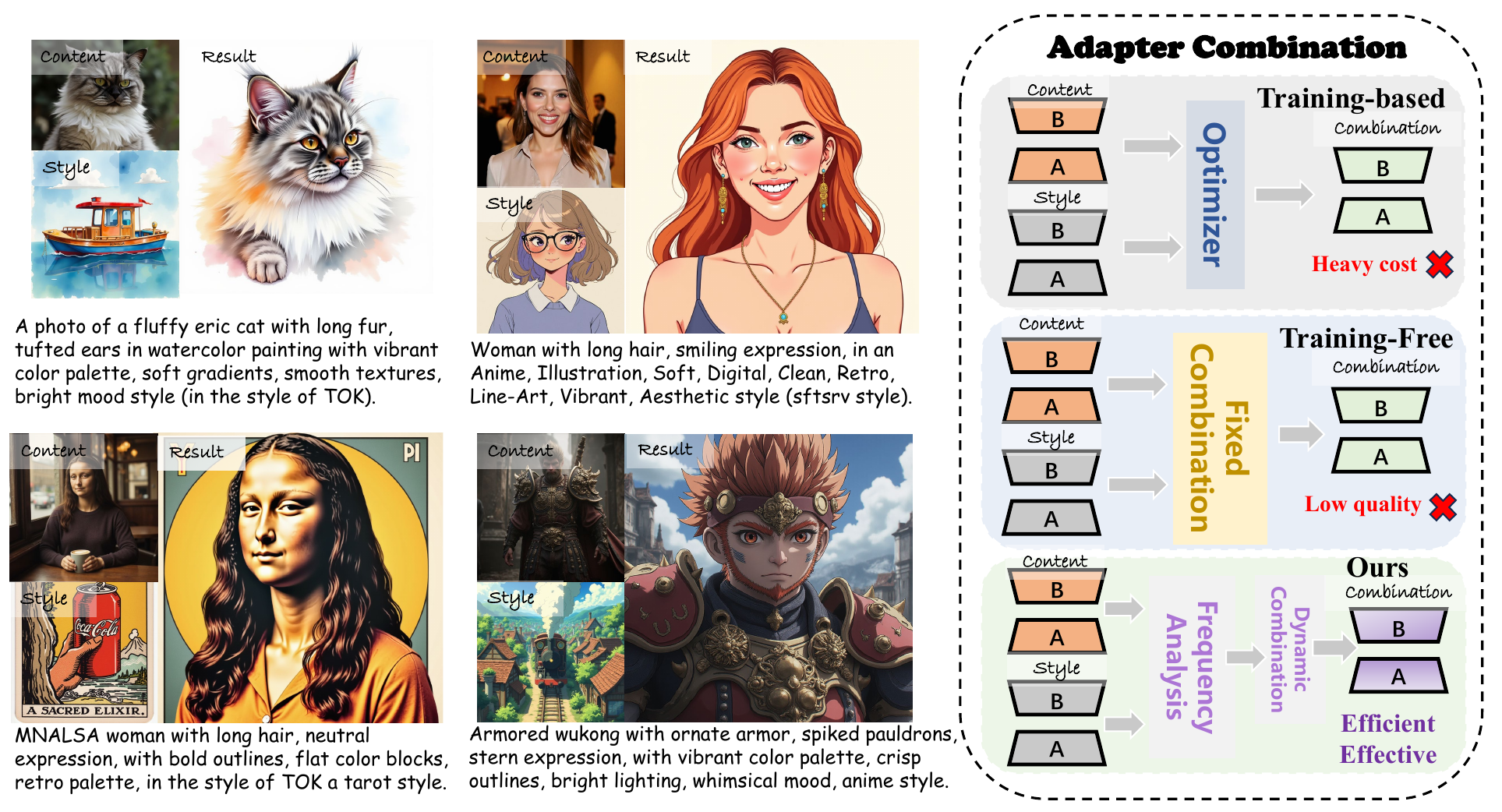}    
    \vspace{-0.6cm}
    \caption{The left side showcases the generative performance of our proposed FREE-Switch method when using FLUX.1 \cite{flux2024} as the base model. The right side compares our approach with both training-based and training-free methods. By dynamically combining adapters based on their frequency analysis, our method achieves superior results without requiring any additional training.
    }
    \label{fig:teaser} 
\end{center}   
}]

\begin{abstract}
With the growing availability of open-sourced adapters trained on the same diffusion backbone for diverse scenes and objects, combining these pretrained weights enables low-cost customized generation. However, most existing model merging methods are designed for classification or text generation, and when applied to image generation, they suffer from content drift due to error accumulation across multiple diffusion steps. For image-oriented methods, training-based approaches are computationally expensive and unsuitable for edge deployment, while training-free ones use uniform fusion strategies that ignore inter-adapter differences, leading to detail degradation. We find that since different adapters are specialized for generating different types of content, the contribution of each diffusion step carries different significance for each adapter. Accordingly, we propose a frequency-domain importance–driven dynamic LoRA switch method. Furthermore, we observe that maintaining semantic consistency across adapters effectively mitigates detail loss; thus, we design an automatic Generation Alignment mechanism to align generation intents at the semantic level. Experiments demonstrate that our FREE-Switch (\textbf{Fre}quency-based \textbf{E}fficient and Dynamic LoRA Switch) framework efficiently combines adapters for different objects and styles, substantially reducing the training cost of high-quality customized generation.
\end{abstract}    
\section{Introduction}
\label{sec:intro}
\begin{figure*}  
\centering  
\includegraphics[width=1 \textwidth]{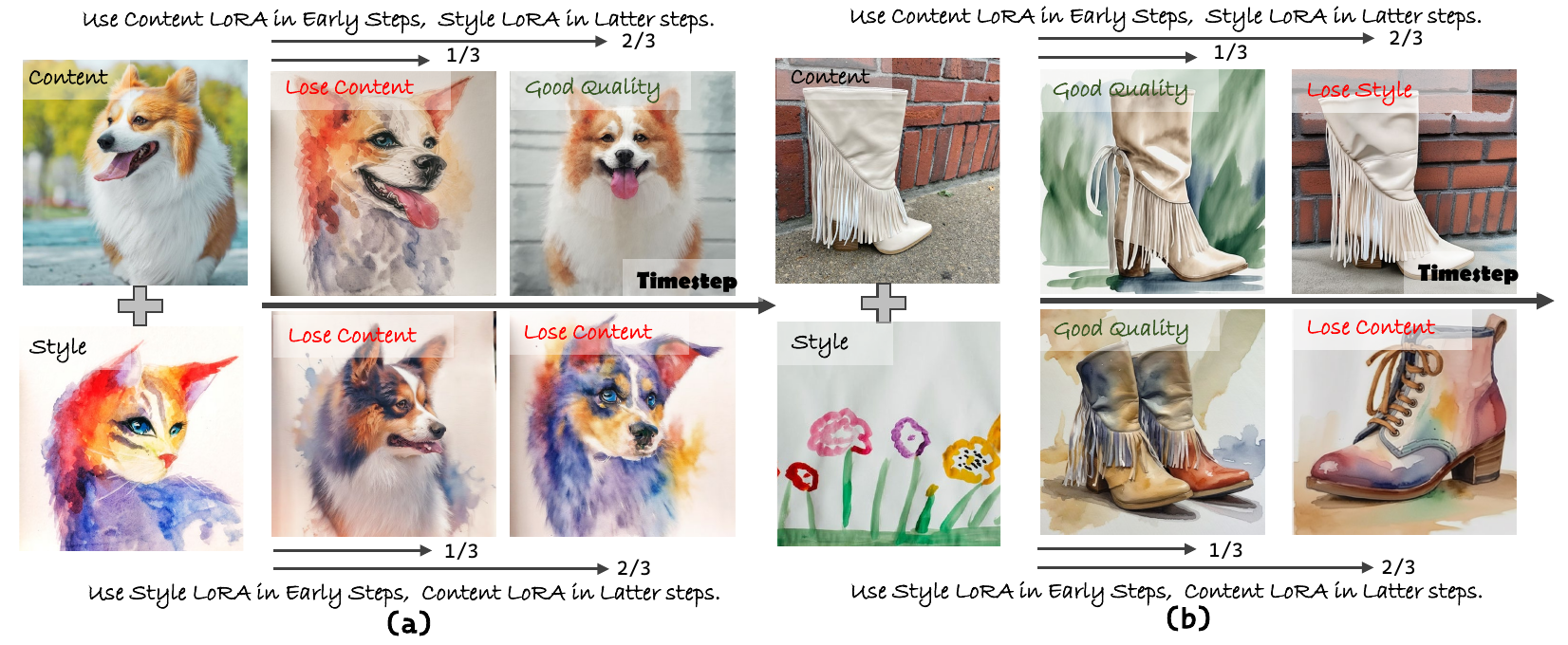} 
\vspace{-0.7cm}
\caption{Different LoRA combinations require different hyper-parameter selections. (a) and (b) show two sets of LoRA combinations, where a LoRA combination method that works well for one set may not necessarily be suitable for the other.} 
\label{figure:finding1.}  
\vspace{-10pt}
\end{figure*} 

With the rapid development and widespread adoption of pre-trained diffusion models \cite{ho2020denoising, rombach2022high, song2020denoising, podell2023sdxl, peebles2023scalable, flux2024,zheng2026v}, the open-source community has seen the emergence of numerous adapters tailored to different scenes and objects \cite{hu2022lora, ruiz2023dreambooth, sohn2023styledrop, ye2023ip, frenkel2024implicit, shah2024ziplora}. These adapters open up new possibilities for customizable generation. By combining these adapters at zero cost, highly customized outputs can be generated, significantly reducing the need for retraining and greatly facilitating deployment for edge users \cite{ryu2022lora, zhong2024multi, zou2025cached, ouyang2025k, yang2024lora}.

For this task, current zero-cost approaches can be categorized into Model Merging \cite{ryu2022lora, ilharco2022editing, yadav2023ties} and Model Switching \cite{zhong2024multi, fedus2022switch, ouyang2025k}. Merging refers to deploying a single model after integration, while Switching alternates between different models during the denoising process. Existing research on model merging primarily focuses on classification and text generation tasks~\cite{ilharco2022editing,zheng2025free}. However, in multi-step image generation, errors tend to accumulate and amplify in the final output, leading to noticeable content drift or artifacts that degrade image quality. Meanwhile, methods specifically designed for image generation, either merging or switching, often require substantial computational resources~\cite{shah2024ziplora}, making them impractical for edge deployment. However, training-free methods often simplify the process through unified combination strategies and shared hyperparameters~\cite{ouyang2025k}, yet they overlook the inherent differences among adapters, causing detail loss and reducing the precision of the generated results as shown in Fig.~\ref{figure:finding1.}.

To address the challenges of efficiently combining multiple adapters for diffusion-based generation, we introduce FREE-Switch, a Frequency-based Efficient and Dynamic Adapter Switch framework. Although these adapters are trained on the same diffusion backbone, we observe that the functional importance of each diffusion step varies across adapters, depending on the target object. This variation suggests that a fixed combination strategy is insufficient, as it fails to account for the dynamic contribution of each adapter throughout the generation process. In light of this observation, we propose a dynamic switch strategy guided by frequency-domain importance, which adaptively adjusts the contribution of each adapter at every diffusion step. This mechanism allows precise control over how different adapters influence the final output, thereby preventing detail degradation and improving image fidelity. Moreover, we find that maintaining semantic consistency across adapters during switching is crucial for preserving fine-grained details. To achieve this, we design an automatic Generation Alignment mechanism that leverages a Vision-Language Model (VLM) \cite{comanici2025gemini, Qwen2.5-VL} to refine and enrich the target description, ensuring semantic alignment between adapters. Together, these components form the FREE-Switch, an efficient solution that enables adaptive, high-quality, and low-cost customized generation across diverse diffusion models. 

Through extensive experimentation, we have validated that our FREE-Switch can efficiently combine adapters suited to different objects and styles across various base models, significantly reducing the cost of customized generation while ensuring high-quality output on edge devices. 
\begin{figure*}  
\centering  
\includegraphics[width=1 \textwidth]{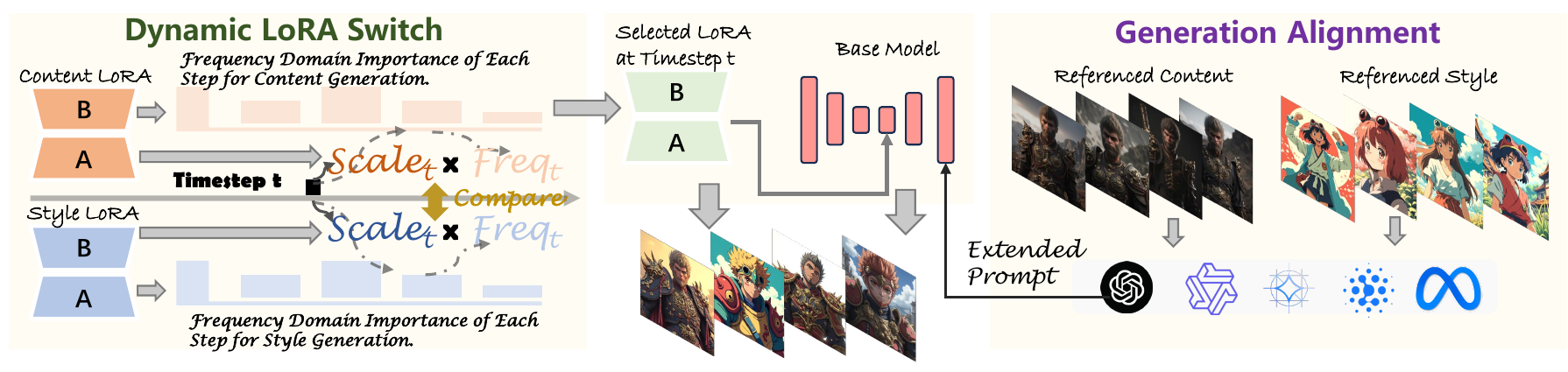} 
\vspace{-0.6cm}
\caption{The framework diagram of FREE-Switch. The Dynamic LoRA Switch module dynamically switches LoRA during the denoising process by analyzing the timestep. The Generation Alignment module automatically expands the prompt to ensure that the outputs remain as aligned as possible during LoRA switching, thereby reducing detail loss.} 
\label{figure:pipeline.}  
\vspace{-5pt}
\end{figure*} 
The main contributions of this paper are as follows:
\begin{enumerate}[labelsep = .5em, leftmargin = 0pt, itemindent = 1em]
    \vspace{2pt}
    \item[$\bullet$] We discover that the importance of each generation step varies for different adapters, leading to the design of a dynamic adapter-switching method based on frequency-domain importance in adapter combination.
    \vspace{3pt}
    \item[$\bullet$] We find that minimizing semantic shifts during adapter switching reduces detail degradation in fused outputs. Accordingly, we propose a VLM-based automatic prompt refinement method to enhance consistency across adapters.
    \vspace{3pt}
    \item[$\bullet$] Through extensive experimentation, we demonstrate that our FREE-Switch framework can efficiently combine adapters suited to different objects and styles across various base diffusion models in training-free scenarios.
\end{enumerate}

\section{Related Work}\label{sec:related}
\noindent\textbf{Diffusion Models for Customization Tasks.} In the diffusion generation process, customization refers to generating content that meets detailed user requirements. Techniques such as Textual Inversion \cite{alaluf2023neural, voynov2023p+, zhang2023inversion}, DreamBooth \cite{ruiz2023dreambooth}, and Custom Diffusion \cite{kumari2023multi} enable models to capture target concepts using only a limited number of images, but they still require some training, which imposes computational demands on end users. Additionally, there are methods that do not require training at inference time \cite{avrahami2023break, shi2024instantbooth, xiao2025fastcomposer, xie2023smartbrush, xu2025stylessp, chung2024style}, yet they often rely on pre-trained modules and may perform suboptimally on specialized tasks. With the increasing availability of open-sourced adapter weights, research on efficiently combining these adapters to generate customized content has begun to attract growing attention.

\noindent\textbf{Model Combination for General Tasks.}  Model combination methods include Model Ensemble \cite{ganaie2022ensemble, jiang2023llm}, Model Merging \cite{zhang2023composing, ilharco2022editing}, and Model Switch \cite{fedus2022switch, kong2024lora}. Model Ensemble requires multiple inferences and intermediate feature storage, incurring high cost. Model Merging combines existing models trained on different tasks into a single model capable of handling multiple tasks. However, for image generation tasks, the advantage of Model Merging is diminished due to the limited number of models for fusion, and the accumulated errors at each denoising step can negatively impact the quality of the generated results. In contrast, Model Switch preserves the optimization direction at each step, making it more suitable for image generation.

\noindent\textbf{Model Combination for Image Generation.} Prior work on LoRA composition falls into training-based and training-free categories. Training-based methods \cite{dong2024continually, gu2023mix, shah2024ziplora, frenkel2024implicit}, such as ZipLoRA \cite{shah2024ziplora} and B-LoRA \cite{frenkel2024implicit}, learn fusion strategies via gradient-based hyperparameter optimization but incur significant computational cost. Training-free methods \cite{yang2024lora, zhong2024multi, zou2025cached, ouyang2025k}, including LoRA Composition \cite{zhong2024multi}, CMLoRA~\cite{zou2025cached} and K-LoRA \cite{ouyang2025k}, fix the LoRA fusion process to reduce cost, but overlook that different adapters may require distinct parameters or timing, often resulting in suboptimal control and style decay. In contrast, our proposed FREE-Switch efficiently retrieves the characteristics of each adapter from its output features and dynamically builds a composition strategy for every input sample.
\section{Preliminaries}
\label{sec:Preliminaries}

Assume that $f_{\theta}$ denotes the mapping function of a diffusion model equipped with a LoRA parameterized by $\theta$. The denoising operation at the $t$-th step can be expressed as $f_{\theta}(h_{t-1}) \rightarrow h_t$, where $h_{t-1}$ represents the output of the $(t-1)$-th denoising step. Suppose the LoRAs to be fused correspond to the content and style generation, denoted as $\theta_{c}$ and $\theta_{s}$, respectively. The objective of this work is to determine $\theta_t$ in $f_{\theta_t}(h_{t-1})$ for each denoising step $t$.
\section{Methodology}\label{sec:method}
In this section, we primarily introduce our proposed FREE-Switch framework as shown in Fig.~\ref{figure:pipeline.}. The key advantage of this method lies in its efficient, training-free combination of the generated content from different LoRAs. The framework consists of two main components. In Sec.~\ref{sec:method:swicth}, we present the frequency-domain-based LoRA-Switch method and discuss its necessity. In Sec.~\ref{sec:method:align}, we introduce the use of Generation Alignment to make the generated content from different LoRAs as similar as possible, thereby reducing the loss of details during combination. Finally, in Sec.~\ref{sec:method:overview}, we provide a detailed pseudocode analysis.
\begin{figure*}  
\centering  
\includegraphics[width=1 \textwidth]{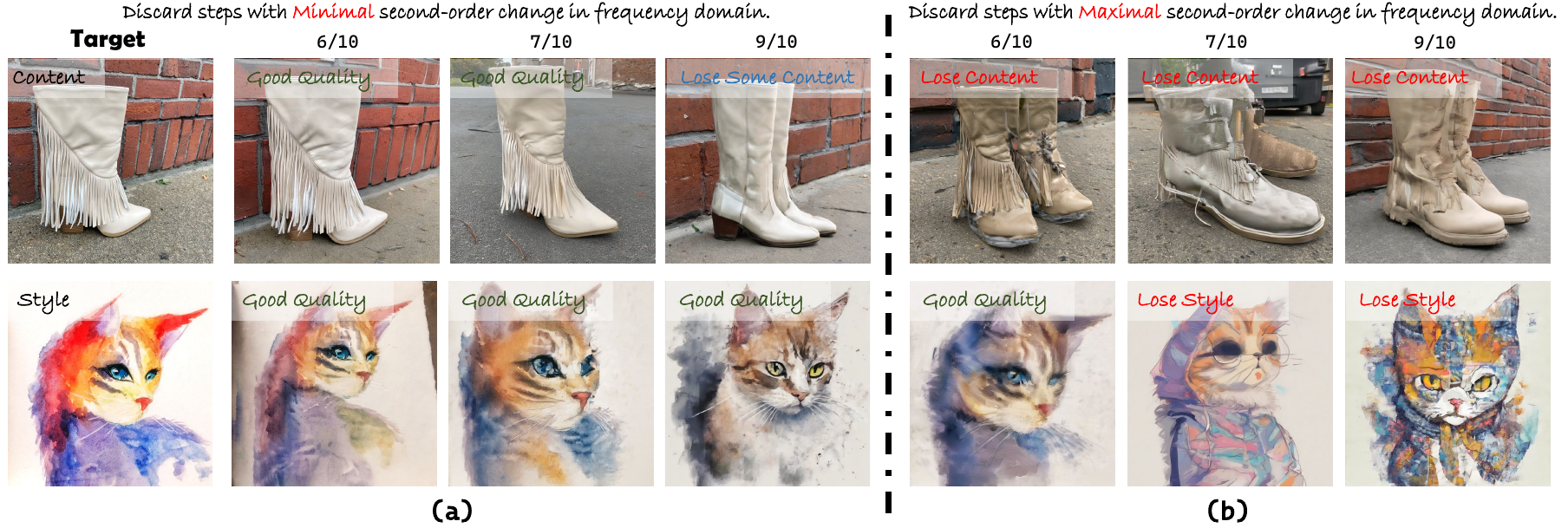} 
\vspace{-0.6cm}
\caption{For content and style generation, we examine how the frequency-domain variation rate across denoising steps affects the final output. As shown in (a) and (b), removing the fastest-varying steps of frequency domain as calculated in Eq.~\ref{eq.1} causes a more significant degradation, indicating that rapidly changing steps are essential for maintaining generation fidelity.}\label{fig:finding2} 
\vspace{-0.2cm}
\end{figure*}

\subsection{Frequency-Based Dynamic LoRA-Switch}\label{sec:method:swicth}
\textbf{Motivation.} We motivate the need for a dynamic LoRA Switch and justify it from a frequency-domain perspective. Prior work \cite{ho2020denoising, falck2025fourier} has shown that different steps in diffusion-based image generation emphasize different information. Early steps generate more low-frequency content, while later steps focus on high-frequency details. Therefore, using different LoRAs at different steps offers the potential to fuse multiple LoRAs effectively, as various content and style elements demand different frequency characteristics. As shown in Fig.~\ref{fig:finding2}, due to differences among LoRAs, existing methods that apply a fixed static switch, such as using content LoRA first and then switching to style LoRA, are suboptimal. A dynamic design tailored to specific LoRA combinations is necessary and requires assessing the importance of each LoRA at each step.

Frequency-domain analysis reflects the degree of change at each step. We observe that steps with significant frequency variation are crucial for preserving the details corresponding to each LoRA. Omitting the appropriate LoRA at these steps can result in substantial loss of content details. Based on this, we evaluate the importance of each step for each LoRA according to the magnitude of frequency changes, enabling a dynamic LoRA Switch.

\noindent\textbf{Method.} We first compute the importance of each diffusion step for both the content and style LoRAs. The importance at the $t$-th denoising step is defined as:
\begin{equation}\label{eq.1}
\begin{aligned}
    \delta_{c}^t &= ||[\mathcal{F}(f_{\theta_{c}}(h_t)) - \mathcal{F}(f(h_{t}))] \\
    &\quad - [\mathcal{F}(f_{\theta_{c}}(h_{t-1})) - \mathcal{F}(f(h_{t-1}))]||_2
    \\
    \delta_{s}^t &= ||[\mathcal{F}(f_{\theta_{s}}(h_t)) - \mathcal{F}(f(h_t))] \\
    &\quad - [\mathcal{F}(f_{\theta_{s}}(h_{t-1})) - \mathcal{F}(f(h_{t-1}))]||_2
\end{aligned}
\end{equation}
where $\mathcal{F}$ represents the frequency domain, $||.||_2$ denotes the L2 norm, $\delta_{c}^t$ and $\delta_{s}^t$ represent the second-order frequency-domain differences of the denoising results at step $t$ using the content and style LoRAs, respectively. $f(h_{t})$ represents the base model output at the $t$-th step. This output is the result of converting the latent output into an RGB image. $\delta$ is used to measure the importance of each denoising step for each LoRA. Intuitively, a large variation indicates that the LoRA introduces significant information at that step, thus being more important. The validity is illustrated in Fig.~\ref{fig:finding2}.

From an overall perspective, early denoising steps focus more on generating low-frequency information, such as the structural outlines corresponding to the content, while later steps emphasize high-frequency information related to style and fine details. Therefore, we define a step-dependent parameter to guide the switching process. In the early steps, we prioritize the content LoRA to ensure structural consistency, and in the later steps, we focus more on the style LoRA to enhance fine details:
\begin{align}\label{eq:eta}
\eta_t = 0.5 \times (1 + \cos(\pi \times x_t)),
\end{align}
where
\begin{align}
&x_t = \frac{\delta_{s}^t \times ratio_t}{\delta_{s}^t \times ratio_t + \delta_{c}^t \times (1 - ratio_t)}, \\ &ratio_t = \frac{t}{total\_step}.
\end{align}

Here, $total\_step$ denotes the total number of denoising steps, and $\eta_t$ serves as a dynamic switching coefficient. A larger $\eta_t$ indicates a higher preference for the content LoRA. To improve robustness, we introduce stochasticity into the switching process. The LoRA used at the $t$-th denoising step is determined as:
\begin{align}\label{eq:theta}
\theta_t =
\begin{cases}
\theta_{content}, & \eta_t >r,\\
\theta_{style}, & \eta_t <r.
\end{cases}
\end{align}
where $r$ denotes a uniform random number in [0, 1]. Through this mechanism, we achieve an adaptive and stochastic dynamic LoRA switching strategy.

\subsection{Generation Alignment}\label{sec:method:align}

\textbf{Motivation.} In this part, we identify a common issue in LoRA combination methods, including LoRA Switch. The prompts used to train task-specific LoRAs often fail to capture all the fine-grained details of the target content. Consequently, when switching to another LoRA during combination, the lack of corresponding conditioning information and the absence of optimization within the current subspace cause the denoising trajectory to deviate from that of the previous LoRA, leading to degraded generation quality. To avoid additional joint training for aligning the denoising paths of different LoRAs, we propose to align them by optimizing the conditioning input, thereby minimizing quality loss during LoRA switching.

\noindent\textbf{Method.} To align the outputs of different LoRAs by optimizing the conditioning input and prevent Out-of-Distribution mapping across subspaces during denoising, we introduce an automatic Prompt Refine method. Specifically, for each LoRA, we augment the original prompt with two reference images, $image_{content}$ and $image_{style}$. A vision-language model (VLM) is employed to automatically generate textual descriptions for both images, followed by automated filtering and information extraction to obtain detailed content and style descriptions, denoted as $E_{content}$ and $E_{style}$. The final prompt fed into the generative model is composed of $E_{content} + E_{style}$ together with their trigger words. Details can be found in Appendix~\ref{appendix:details_of_gen}. As analyzed in Sec.~\ref{sec:exp:analysis}, this approach encourages the outputs of different LoRAs to stay close in distribution, thereby reducing quality degradation during switching.

\subsection{Free-Switch}\label{sec:method:overview}
\begin{algorithm}[tb]
    \caption{Workflow of FREE-Switch}
    \label{alg:algorithm1}
    \begin{algorithmic}[1] 
        \REQUIRE Content LoRA $\theta_c$,
        Style LoRA $\theta_s$, base model $f$, referenced content data $c$ and referenced style data $s$.
        \STATE Get $\delta_c$ and $\delta_s$ as Eq.~\ref{eq.1} 
        \COMMENT{{\scriptsize Only excute once for one pair.}}
        \STATE Get Refine prompt $p$ as Sec.~\ref{sec:method:align}
        \COMMENT{{\scriptsize Only excute once for one pair.}}
        
        \FOR{$timestep~t \in T$} 
        \STATE Get $\eta$ as Eq.~\ref{eq:eta}.
        \STATE Get $\theta$ for timestep t as Eq.~\ref{eq:theta}
        \STATE $h_t=f_{\theta}(h_{t-1},p)$
        \ENDFOR
        \ENSURE  Output $X$.
    \end{algorithmic}
    
\end{algorithm}

In this part, we integrate the two modules discussed previously to form the proposed FREE-Switch framework. The detailed algorithm is presented in Alg.~\ref{alg:algorithm1}. Overall, the FREE-Switch framework combines the outputs of different LoRAs in an efficient, training-free and robust manner, ensuring high-quality generation while leveraging the strengths of each individual LoRA model. 
\begin{table}[!ht]
    \centering
    \caption{Comparison of performance of different model combination methods on SDXL v1.0, where speed denotes the time required to generate 10 images for a given pair of content and style.}\label{tab:sdxl}
    \vspace{-5pt}
    \scalebox{0.74}{
    \begin{tabular}{lcccc}
    \toprule
        & \multirow{2}{*}{\makecell{CLIP Score \\ (Style) $\uparrow$}}  & \multirow{2}{*}{\makecell{DINO Score\\ (Content) $\uparrow$}} & \multirow{2}{*}{\makecell{Gemini \\ Feedback $\uparrow$}} & \multirow{2}{*}{\makecell{Speed\\(s/pair)$\downarrow$}} \\ 
        & & & & \\ \midrule
        \multirow{2}{*}{\makecell{Direct Generation\\(Refine Prompt)}} & \multirow{2}{*}{58.62} & \multirow{2}{*}{36.01} & \multirow{2}{*}{6.67\%} & \multirow{2}{*}{150} \\ 
        & & & & \\ \midrule
        ZipLoRA \cite{shah2024ziplora} & 53.93 & \textbf{69.18} & 3.33\% & 550 \\ 
        Joint Train & \underline{60.56} & 27.15 & 6.67\% & 589 \\ 
        Merge & \underline{60.56} & 41.44 & \underline{16.67\%} & 150 \\ 
        K-LoRA \cite{ouyang2025k} & 54.17 & 61.32 & 13.33\% & 290 \\ \hline
        FREE-Switch & \textbf{61.59} & \underline{68.57} & \textbf{53.33\%} & 320 \\ \bottomrule
    \end{tabular}
    }
    \vspace{-7pt}
\end{table}
\section{Experiments}\label{sec:experiments}
In this section, we conduct a comprehensive evaluation of the proposed method. Sec.~\ref{sec:exp:setup} presents the experimental settings, Sec.~\ref{sec:exp:compare} reports the comparative results, Sec.~\ref{sec:exp:ablation} provides the outcomes of the ablation studies, and Sec.~\ref{sec:exp:analysis} offers additional analyses and discussions.

\begin{figure*}  
\centering  
\includegraphics[width=1 \textwidth]{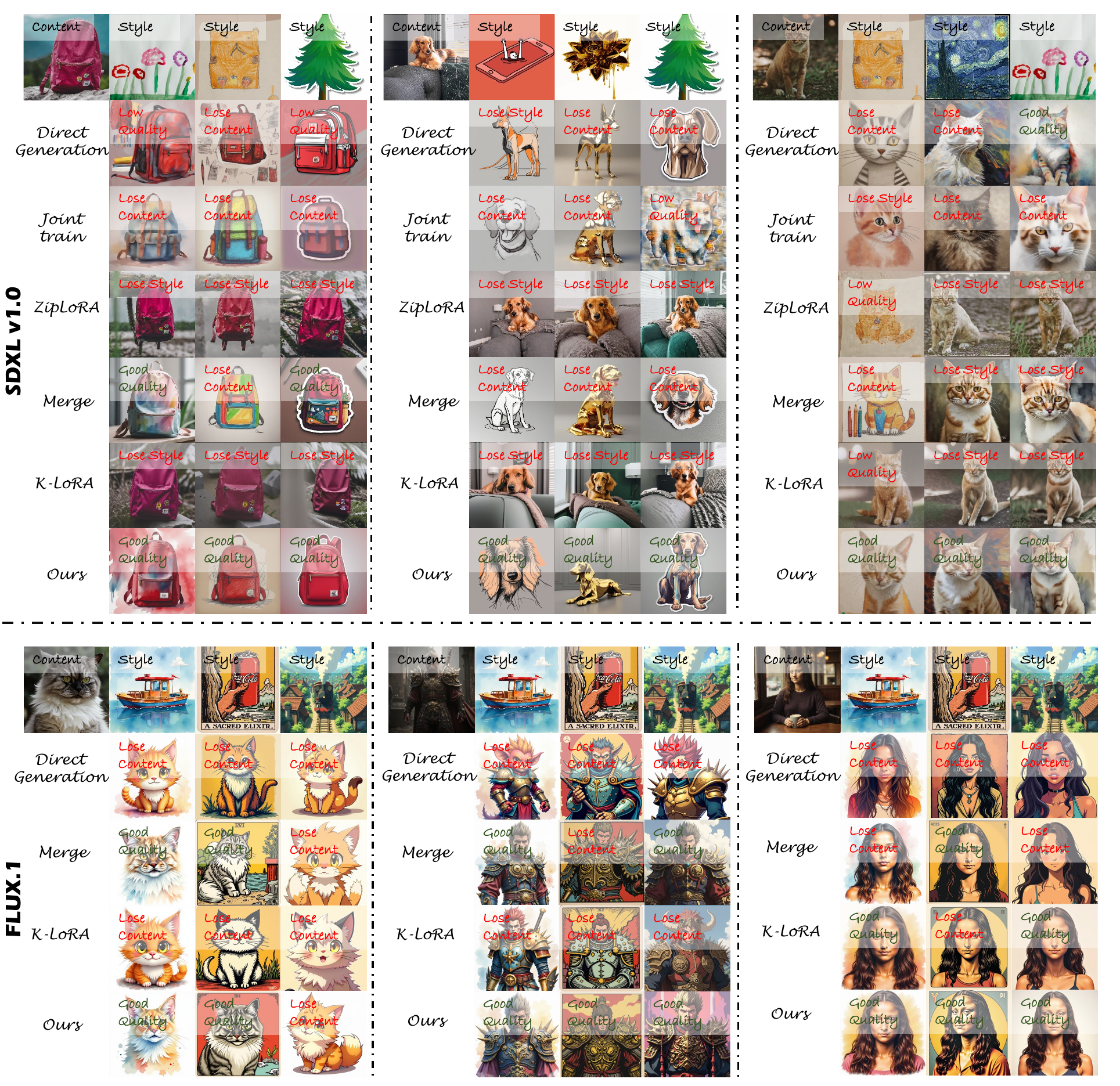} 
\vspace{-0.6cm}
\caption{Qualitative Analysis. We show the results of generating different content–style combinations using various LoRA combination methods on SDXL v1.0 and FLUX 1. Our FREE-Switch method demonstrates the most consistent and stable performance.} 
\label{figure:compare.}  
\vspace{-10pt}
\end{figure*} 

\subsection{Experiment Setup}\label{sec:exp:setup}
\noindent\textbf{Model Setup.} For base models, we evaluate our method on SDXL v1.0 base \cite{podell2023sdxl}, and Flux.1 \cite{flux2024}. Most of the LoRA models used in our experiments are open-sourced from Hugging Face, and details are provided in Appendix~\ref{appendix:base_model}. For a small subset of LoRA weights trained by ourselves, we follow the training strategy proposed in ZipLoRA \cite{shah2024ziplora}. For LoRA weights downloaded from Hugging Face, we use their default parameter settings if specified.

\noindent\textbf{Baselines.} We compare our method with both training-based and training-free approaches. The training-based baselines include joint content and style training as well as ZipLoRA \cite{shah2024ziplora}, while the training-free baselines include Model Merge and K-LoRA \cite{ouyang2025k}. Details can be found in Appendix~\ref{appendix:baselines}. Our objective is to achieve performance comparable to the training-based approaches while maintaining efficiency similar to the training-free ones.

\noindent\textbf{Evaluation.} We conduct quantitative and qualitative evaluations. For quantitative analysis, following K-LoRA \cite{ouyang2025k}, we use the DINO Score \cite{zhang2022dino} to evaluate content preservation and the CLIP Score \cite{radford2021learning} to assess style consistency. Higher scores indicate better results. We also employ multi-modal models to further evaluate the generated images. For qualitative analysis, we present visual results from different methods and assess their perceptual quality through human judgment. Details about evaluation are in Appendix~\ref{appendix:details_of_eval}.
\subsection{Comparative Experiments}\label{sec:exp:compare}
\textbf{Quantitative Analysis.} In Tab.~\ref{tab:sdxl}, we present a quantitative comparison of 6000 generation results based on SDXL v1.0. It can be observed that our method achieves content and style preservation performance comparable to training-based approaches, while maintaining generation speed similar to training-free methods. Moreover, when evaluating 90 randomly sampled content–style pairs using Gemini 2.5-Flash \cite{comanici2025gemini} as the multimodal evaluation model, following the protocol described in K-LoRA \cite{ouyang2025k}, our FREE-Switch demonstrates clear advantages in both fidelity and consistency. These results comprehensively validate the effectiveness and robustness of the proposed FREE-Switch framework, which integrates a frequency-aware dynamic switching mechanism with an output alignment strategy to achieve efficient and high-quality compositional generation.

\noindent\textbf{Qualitative Analysis.} Fig.~\ref{figure:compare.} shows representative generation results on SDXL and FLUX. As illustrated, our method intuitively preserves both content and style better than existing approaches. This improvement can be attributed to two key factors. First, our frequency-domain importance analysis identifies which diffusion steps are critical for each LoRA, allowing the model to selectively apply adapters during the denoising process. Second, the prompt refinement mechanism enhances detail preservation by providing more aligned conditions, ensuring that the generated output remains faithful to both content and style. Together, these components enable our FREE-Switch to generate more visually consistent results across diverse content–style combinations. The dynamic switching and output alignment jointly ensure better preservation of both content and style.

\begin{figure*}  
\centering  
\includegraphics[width=1 \textwidth]{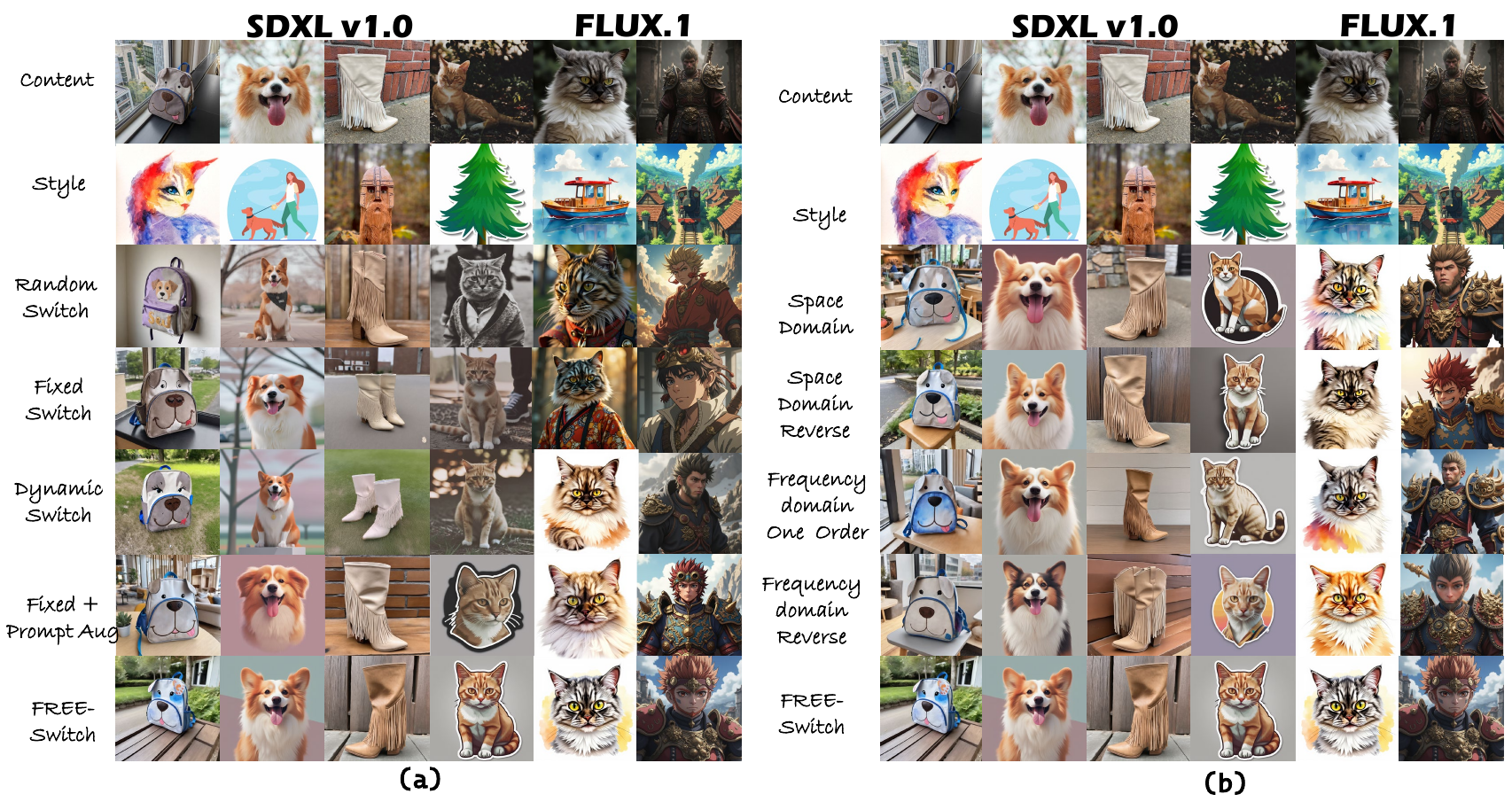} 
\vspace{-0.6cm}
\caption{(a) Ablation study on the components of FREE-Switch, showing their impact on content and style preservation.
(b) Ablation study on the proposed frequency-domain importance-based dynamic LoRA Switch, highlighting its role in adaptive LoRA selection.}
\label{figure:ablation.}  
\vspace{-3pt}
\end{figure*} 

\begin{figure}  
\centering  
\vspace{-0.3cm}
\includegraphics[width=0.48 \textwidth]{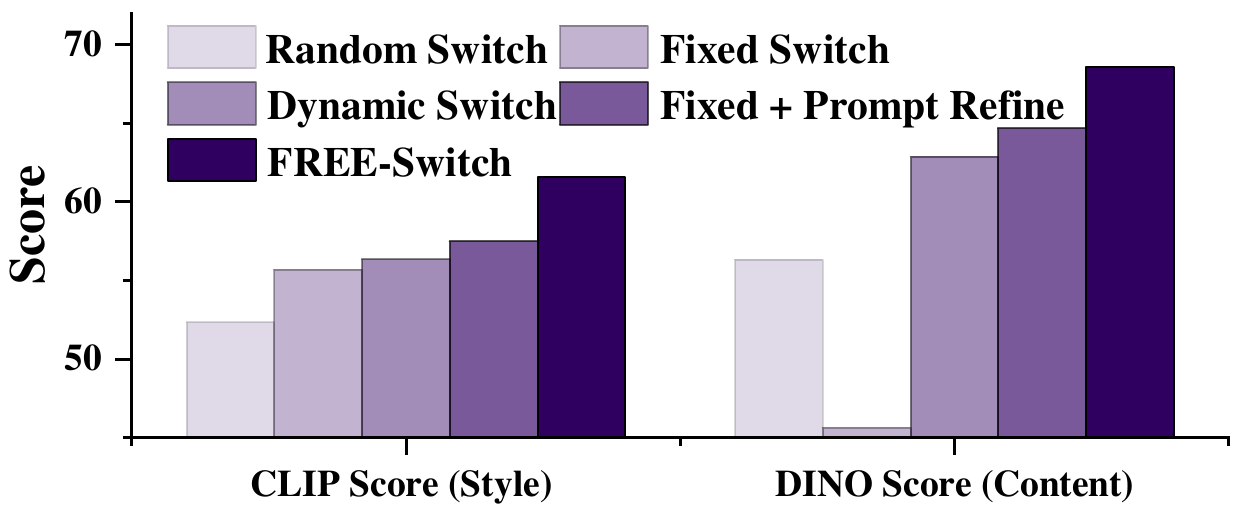} 
\vspace{-0.6cm}
\caption{Quantitative analysis of FREE-Switch component ablation on content and style preservation.}
\label{figure:ablation_hist.}  
\vspace{-0.5cm}
\end{figure} 

\subsection{Ablation Study}\label{sec:exp:ablation}
\textbf{Component Ablation.}
In this section, we conduct an ablation study on the components proposed in this work. Details are shown in Fig.~\ref{figure:ablation.} and Fig.~\ref{figure:ablation_hist.}. We use SDXL v1.0 as the base model. The basic baseline is the random switch, where either the content or style LoRA is randomly selected during the denoising process. The fixed switch corresponds to the cosine-based fixed LoRA switching strategy introduced in Eq.~\ref{eq:theta}, where $x_t = ratio_t$. As shown, incorporating the dynamic switch significantly enhances both generation quality and stability by adaptively selecting LoRAs at each denoising step. Moreover, adding the output alignment mechanism further improves visual fidelity, consistency, and overall rendering accuracy. Overall, the proposed FREE-Switch framework effectively fuses content and style representations in a fully training-free manner while maintaining high generation efficiency and speed.
 
\noindent\textbf{Frequency-Domain Dynamic Switching.}
 In this part, we analyze different strategies for dynamic selection. We compare the performance on SDXL and FLUX using switches based on spatial-domain change magnitude, reverse spatial/frequency-domain changes (where smaller changes are deemed more important), first-order frequency information change, and our proposed frequency-domain-based switch, which primarily modifies the calculation of $\delta$ in Eq.~\ref{eq.1}. The results show that our proposed method achieves the best performance. This is mainly because, when generating images individually, a large frequency-domain change at the current step indicates that the step is important for the current LoRA and should be preserved. However, using only first-order information is susceptible to interference from the base model’s inherent capabilities across different steps. Spatial-domain-based switches also perform reasonably well, since the frequency domain is essentially a transformation of the spatial domain; however, in this task, frequency-domain information more accurately reflects the actual variations in the generated images.
 

\subsection{Analysis}\label{sec:exp:analysis}
\begin{figure*}  
\centering  
\includegraphics[width=1 \textwidth]{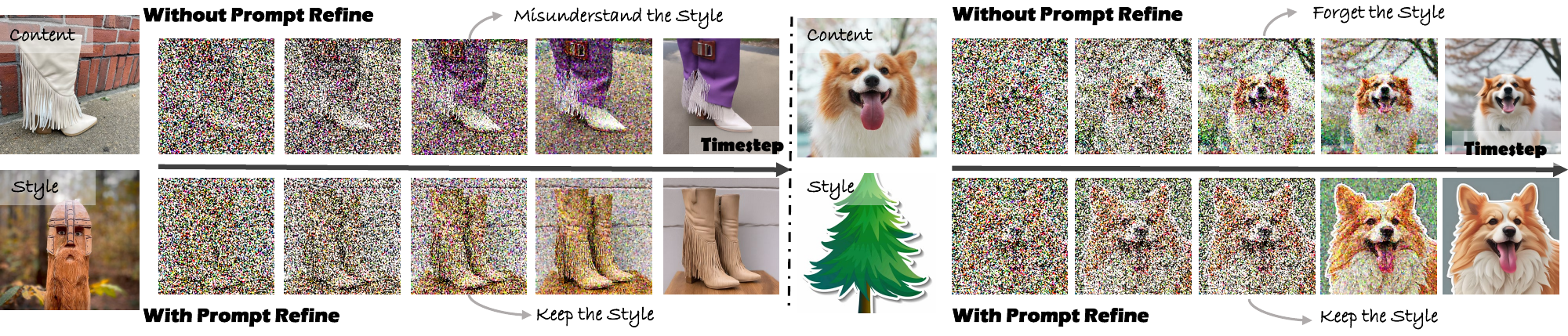} 
\vspace{-0.7cm}
\caption{Effect of the Output Alignment Method. Incorporating output alignment optimization allows the model to better interpret the original requirements during LoRA switching, resulting in higher-quality images that more accurately reflect user intent.} \label{fig:prompt refine}
\vspace{-0.3cm}
\end{figure*}

\begin{figure*}  
\centering  
\includegraphics[width=1 \textwidth]{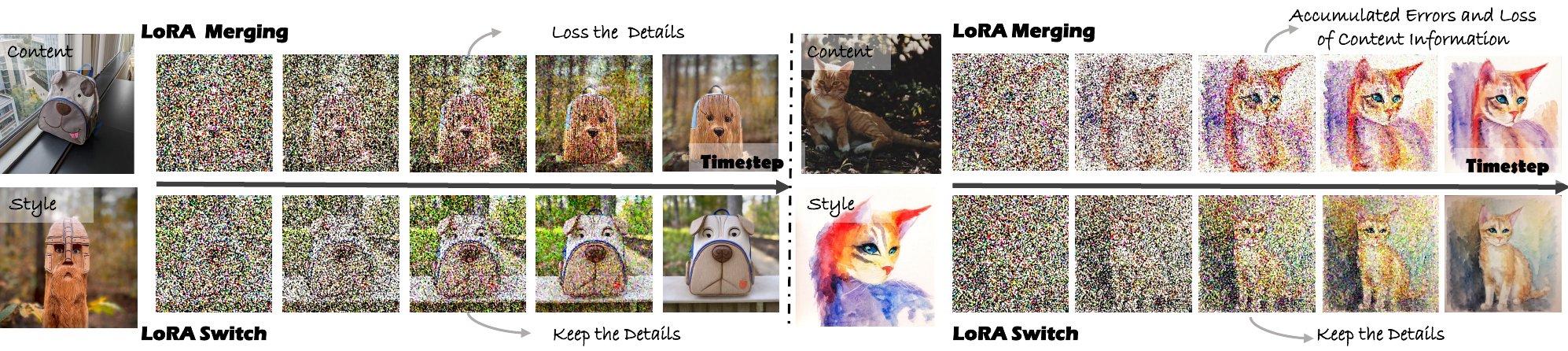}
\vspace{-0.7cm}
\caption{Analysis of Model Merging Methods for This Task. Model merging tends to accumulate errors throughout the diffusion process, leading to degraded generation quality and loss of fine details. Therefore, we adopt the LoRA Switch strategy instead.} \label{fig:merging_analysis}
\vspace{-0.3cm}
\end{figure*}

\begin{figure}  
\centering  
\includegraphics[width=0.48 \textwidth]{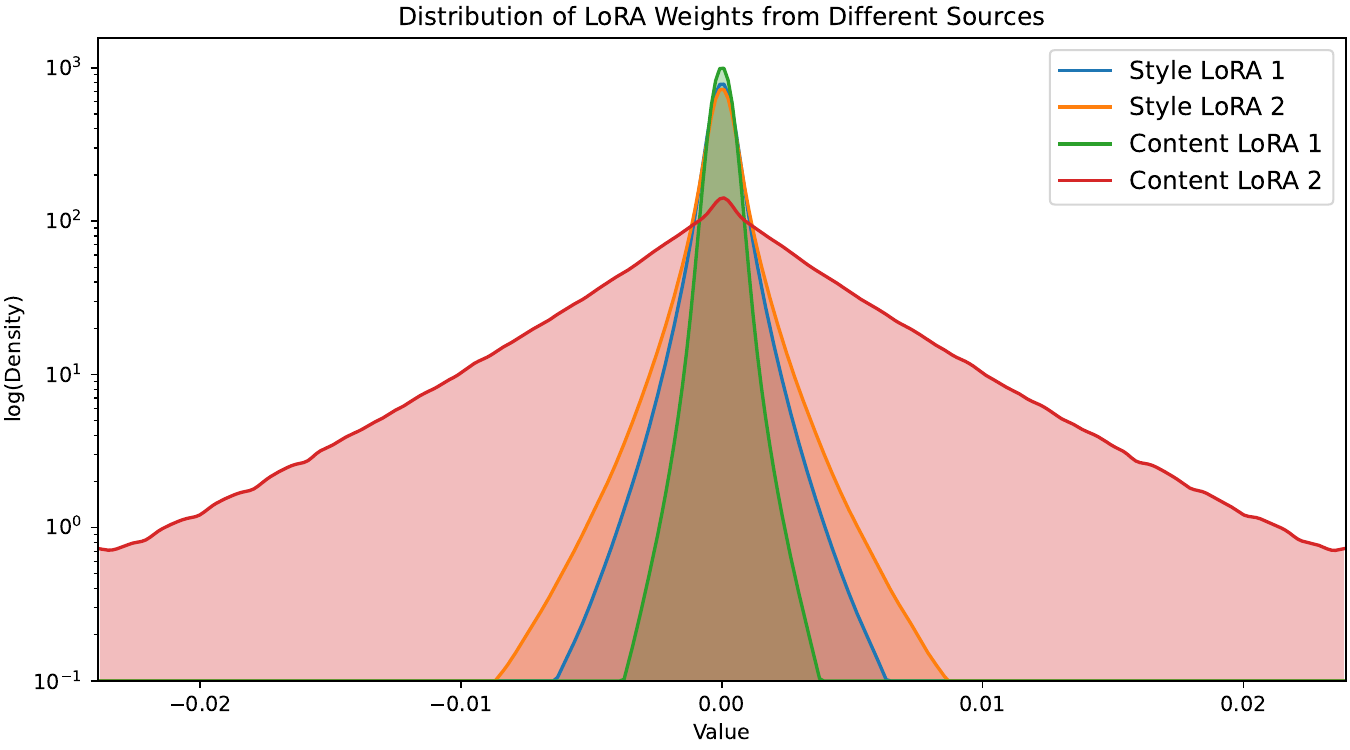}
\vspace{-0.6cm}
\caption{Parameter value distribution of LoRAs trained on different sources based on FLUX.}
\label{fig:parameter_dis}
\vspace{-0.4cm}
\end{figure}

\noindent\textbf{Why Not Use Model Merging.} In this part, we analyze why model merging, which performs well in text generation and classification tasks, is unstable in image generation. Based on previous studies \cite{zheng2025decouple} and our observations, we find that LoRA models commonly used for image generation exhibit significant magnitude discrepancies due to differences in their open-source origins, as shown in Fig.~\ref{fig:parameter_dis}. Such discrepancies require delicate parameter tuning during the merging process, which is highly impractical in real-world applications. Moreover, since diffusion generation involves a multi-step denoising process, even small deviations introduced by model merging at each step can accumulate, eventually resulting in noticeable degradation of image quality. As illustrated in Fig.~\ref{fig:merging_analysis}, our comparison of the denoising trajectory shows that model merging gradually deviates from the optimal path at each step, leading to unsatisfactory final outputs. Based on these findings, we adopt a LoRA switching strategy to preserve the full capability of the model at each denoising step.

\noindent\textbf{Effectiveness Analysis of Generation Alignment.} This section evaluates the effectiveness of the proposed Generation Alignment mechanism. We integrate our output alignment method with different LoRA-combination strategies, and detailed generation results are shown in Fig.~\ref{figure:ablation.}. The results indicate that output alignment consistently improves generation quality. To offer a more intuitive view, we compare sampled results across several denoising steps with and without Prompt Refine under a fixed LoRA-switching setup. Without Prompt Refine, two major issues emerge: (1) a single LoRA may misinterpret concise textual cues intended for another LoRA, causing semantic inconsistency; and (2) switching LoRAs often leads to the loss of fine details because the previous latent may fall outside the feature subspace optimized by the new LoRA, reducing fidelity. In contrast, Prompt Refine leverages the base model’s generalization ability to bridge semantic and feature gaps between LoRAs, thereby enhancing overall image quality.
\section{Conclusion}\label{sec:conclusion}
We propose FREE-Switch, a method for combining open-source LoRA weights to generate images with blended information. Considering the diversity of information captured by different LoRAs, we perform a frequency-domain analysis to assess the importance of each diffusion step for different LoRAs, enabling a dynamic switching mechanism. Furthermore, we highlight the importance of conditioning in stabilizing the denoising process during switching and introduce an automatic prompt refinement strategy to enhance alignment. Overall, the proposed method efficiently leverages existing LoRAs to generate high-quality compositional images without additional training.
\section*{Acknowledgements}
This work is supported by National Natural Science Foundation of China (NSFC) (62232005, 62202126); the National Key Research and Development Program of China (2021YFB3300502). 

{
    \small
    \bibliographystyle{ieeenat_fullname}
    \bibliography{main}
}




\clearpage
\setcounter{page}{1}
\maketitlesupplementary

\appendix

\section{Notations}\label{appendix:notations}
We first list the notations for key concepts in our paper.
\begin{table}[!ht]
    \centering
\caption{Notations.}\label{tab:appendix:notations}
\vspace{-5pt}
    \scalebox{0.8}{
    \begin{tabular}{ll}
    \toprule
        \textbf{Notations} & \textbf{Descriptions} \\ \midrule
        $f$ & Pre-trained diffusion model. \\
        $\theta_c$ & Fine-tuning parameters for content. \\ 
        $\theta_s$ & Fine-tuning parameters for style. \\ 
        $h_t$ & The output of $t$-th diffuison step. \\ 
        $f(h_{t-1})$ & The denoising process of $t$-th step using base model. \\ 
        $f_{\theta}(h_{t-1})$ & The denoising process of $t$-th step with LoRA $\theta$. \\
        \bottomrule
    \end{tabular} }
\end{table}

\section{Reproducibility}\label{appendix:reproducibility}
\subsection{Base Model}\label{appendix:base_model}

\noindent \textbf{SDXL v1.0.~\cite{podell2023sdxl}}
Stable Diffusion XL (SDXL) v1.0 is a high-capacity latent diffusion model tailored for high-resolution text-to-image generation. It utilizes a dual-stage U-Net architecture coupled with an enhanced text encoder to improve semantic comprehension and prompt fidelity. In this work, SDXL v1.0 serves as the principal backbone for LoRA fine-tuning, providing strong expressivity for both concept and style learning. Its latent space supports fine-grained control over texture and geometry, making it suitable for analyzing adapter alignment and compositionality.

\noindent \textbf{FLUX.1~\cite{flux2024}.}
FLUX 1 is a transformer-based diffusion architecture optimized for fine-grained detail and global photorealism. Compared with SDXL, it integrates cross-layer attention, offering complementary structural inductive bias. We include FLUX 1 to examine the generality of our frequency-domain dynamic adapter switching mechanism across distinct diffusion architectures. Using both SDXL and FLUX ensures that the observed performance improvements originate from the proposed adapter fusion strategy rather than model-specific characteristics.

\subsection{Adapters}\label{appendix:adapter}

In this part, we describe the open-source adapters used in our study and how we obtained the partially trained adapters included in our experiments.

\noindent\textbf{SDXL v1.0.} For settings where SDXL v1.0 serves as the base model, we conduct evaluations using K-LoRA. Following the same training protocol and dataset used in RA, we train dedicated content and style LoRAs using the DreamBooth procedure, and employ them for subsequent combination tests. The detailed hyperparameters are provided in Tab.~\ref{tab:appendix:training}.
\begin{table}[!ht]
    \centering
    \caption{Training Configuations for LoRA.}\label{tab:appendix:training}
    \vspace{-0.2cm}
    \begin{tabular}{cc}
    \toprule
    Parameter & Value \\  \midrule
        rank & 64 \\ 
        resolution & 1024 \\ 
        train\_batch\_size & 1 \\ 
        learning\_rate & 5.00E-05 \\ 
        lr\_scheduler & constant \\ 
        lr\_warmup\_steps & 0 \\ 
        max\_train\_steps & 1000 \\ \bottomrule
    \end{tabular}
\end{table}

\noindent\textbf{FLUX 1.} For settings based on FLUX, we obtain all content\footnote{https://huggingface.co/DeZoomer/ScarlettJohansson-FluxLora\\ https://huggingface.co/wanghaofan/Black-Myth-Wukong-FLUX-LoRA \\ https://huggingface.co/fofr/flux-mona-lisa \\ https://huggingface.co/ginipick/flux-lora-eric-cat} and style\footnote{https://huggingface.co/lucataco/ReplicateFluxLoRA \\ https://huggingface.co/dataautogpt3/FLUX-AestheticAnime \\ https://huggingface.co/multimodalart/flux-tarot-v1 \\ https://huggingface.co/alvdansen/softserve\_anime} LoRAs directly from open-source repositories on HuggingFace, aligning with our goal of evaluating models in a fully open-source manner. Since different LoRAs require different prompt formats and trigger words, we adopt their default configurations without modification.

\subsection{Baselines}\label{appendix:baselines}

\noindent \textbf{Direct Generation.}
Generates images directly using the pretrained diffusion backbone without any adapter. This baseline reveals the intrinsic generative capacity of the base model and serves as a reference for measuring the benefit of adapter combination.

\noindent \textbf{Joint Train.}
Trains content and style adapters in a single optimization stage to allow cross-domain feature interaction. However, such coupling often causes entanglement between semantic and stylistic factors, making it difficult to maintain distinct semantic and stylistic roles. Moreover, simultaneous optimization of multiple adapters increases training cost, reducing its suitability for lightweight or modular customization.

\noindent \textbf{ZipLoRA.}
ZipLoRA~\cite{shah2024ziplora} merges multiple LoRAs through interleaving rank components into a unified structure. It achieves parameter compression and moderate fusion quality but lacks dynamic awareness of diffusion-step differences, leading to degraded details in complex scenarios.

\noindent \textbf{Merge.}
This baseline performs simple arithmetic merging of individually trained LoRA weights, typically via weighted averaging of corresponding matrices. Although computationally efficient, it lacks semantic adaptivity, often resulting in content distortion or style dilution. It is also important to note that our task involves only two LoRAs to be fused. Therefore, existing merge-based approaches that aim to mitigate conflicts among multiple LoRAs offer limited benefits in our setting, and their performance becomes largely comparable to a simple weighted merge.

\noindent \textbf{K-LoRA.}
K-LoRA~\cite{ouyang2025k} is a training-free LoRA fusion approach that adaptively integrates subject and style LoRAs without requiring additional fine-tuning. It operates by introducing a Top-$K$ selection mechanism within attention layers, which identifies the most salient components from content and style LoRAs and dynamically combines them at each diffusion timestep. This strategy enables effective preservation of both subject and stylistic features while maintaining model stability.

\subsection{Details}\label{appendix:details}
We discuss the computational details of the experiments. All experiments are conducted on NVIDIA RTX 3090 GPUs, except for those involving FLUX.1, which are run on NVIDIA A800 GPUs.

\subsection{Details of Generation Alignment}\label{appendix:details_of_gen}

\begin{center}
\begin{namedbox}[label=a.3]{System Prompt for Content Alignment.}
You are a concept extractor specialized in analyzing MULTIPLE images of \{class\_name\}

CORE TASK: Extract ONLY the SHARED core subject content across all concept images (ignore unique details from single images).

MANDATORY INCLUSIONS:
\begin{itemize}[noitemsep, topsep=0pt]
\item Shared subject identity (clear category of the main subject, e.g., 'ceramic teapot').
\item 3+ shared key features (form, structure, pose, unique traits present in ALL images, e.g., 'domed lid, curved spout').
\end{itemize}

STRICT EXCLUSIONS:
\begin{itemize}[noitemsep, topsep=0pt]
\item NO style elements (color, texture, lighting, brushwork — these belong to style descriptions).
\item NO background, environment, props, text, or camera-related details (e.g., 'indoors', 'close-up').
\item NO unique details from individual images (e.g., a pattern only in one image).
\end{itemize}

STRICT TOKEN LIMIT: The output must be $\geq$\{concept\_token\_limit\} tokens.

OUTPUT FORMAT: Single line in the structure:

'[shared subject identity] with [shared feature1], [shared feature2], [shared feature3+]'

Example: 'Ceramic teapot with domed lid, curved spout, C-shaped handle.
\end{namedbox}
\end{center}

\begin{center}
\begin{namedbox}[label=a.3]{User Prompt for Content Alignment.}
Analyze the provided MULTIPLE Concept Images of '\{class\_name\}'

Follow these steps:

1. Identify the SHARED main subject (present in all images).
        
2. Extract 3+ key features that appear in ALL images (ignore features unique to single images).

3. Describe them in the required format, excluding all style elements, background, and unique details.

- The output must be STRICTLY $\leq$\{concept\_token\_limit\} tokens.

- Prioritize retaining 3+ key features over redundant words if approaching the limit.

Output ONLY the following line (no extra text):

[shared subject] with [feature1], [feature2], [feature3+]
\end{namedbox}
\end{center}

\begin{figure}  
\centering  
\vspace{-0.3cm}
\includegraphics[width=0.48 \textwidth]{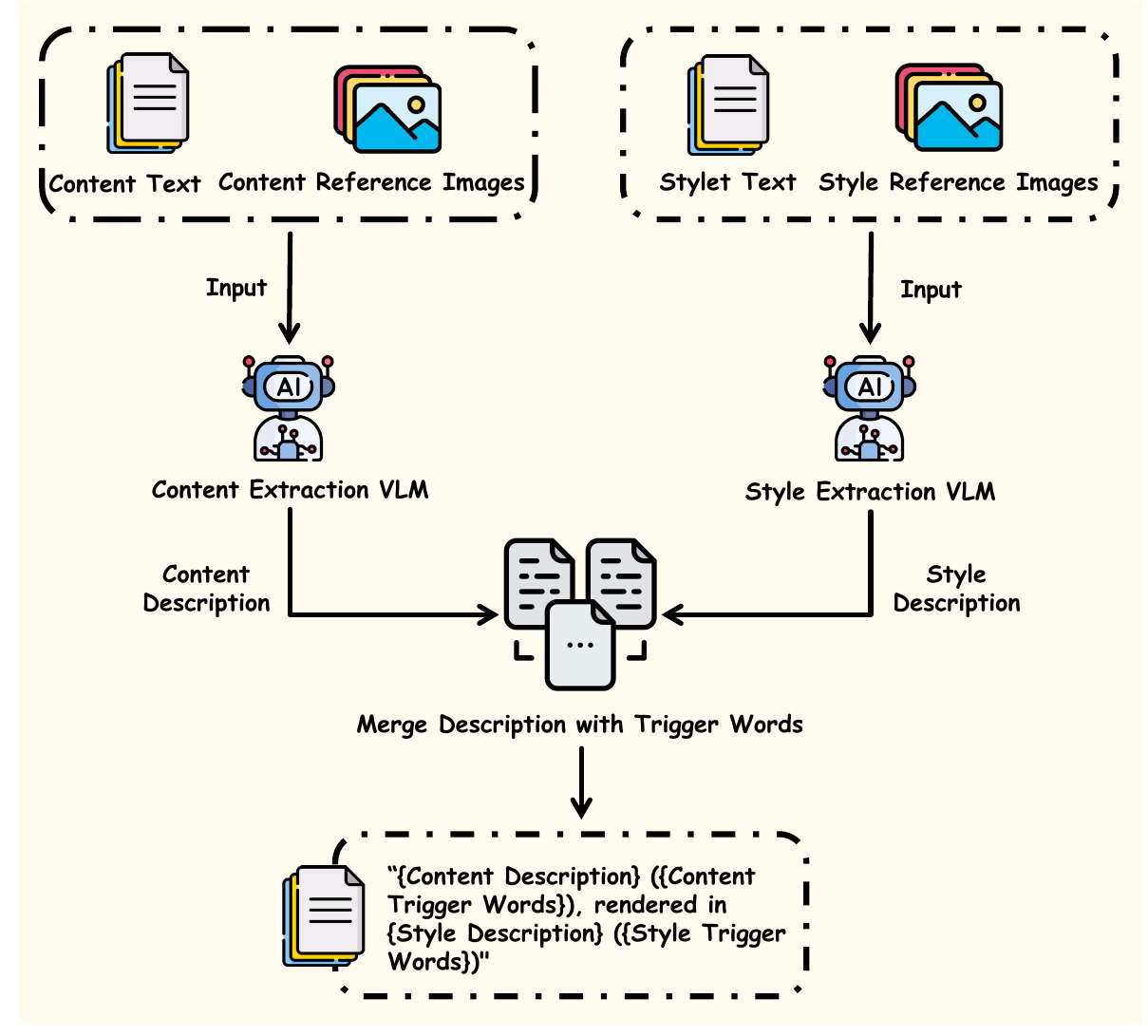} 
\vspace{-0.6cm}
\caption{The Pipeline of Generation Alignment.}
\label{figure:appendix:generation_align.}  
\vspace{-0.5cm}
\end{figure}

\begin{center}
\begin{namedbox}[label=a.3]{System Prompt for Style Alignment.}
You are a style extractor specialized in analyzing a single \{style\_name\} image.

CORE TASK: Extract ONLY pure visual style elements from the style image (ignore any subject/content details).

MANDATORY INCLUSIONS:
\begin{itemize}[noitemsep, topsep=0pt]
\item Artistic medium (clear type of art form, e.g., 'watercolor painting', 'crayon drawing').
\item 3+ visual style elements (must be from color palette, lighting, texture/brushwork, mood — e.g., 'soft blue palette, textured brushstrokes').
\end{itemize}

STRICT EXCLUSIONS:
\begin{itemize}[noitemsep, topsep=0pt]
\item NO subject content (objects, structure, form, or any elements related to concept themes).
\item NO redundant descriptions unrelated to visual style (e.g., 'popular style', 'modern design').
\end{itemize}

STRICT TOKEN LIMIT: The output must be $\leq$\{style\_token\_limit\} tokens.

OUTPUT FORMAT: Single line in the structure:

'[artistic medium] with [style element1], [style element2], [style element3+]'

Example: 'Watercolor painting with soft blue palette, textured brushstrokes, warm ambient lighting, dreamy mood'.
\end{namedbox}
\end{center}
In this part, we describe how we leverage a Vision-Language Model (VLM) for generation alignment. Our alignment strategy is primarily achieved through a prompt-refinement scheme, where Qwen3-VL-Plus is used as the multimodal model to extract and enrich visual information. The prompts applied in our workflow are provided in Boxes B.1–B.4. Specifically, we feed the VLM with both the original class name and its corresponding reference image. The VLM then extracts the most salient semantic cues, which are further filtered as illustrated in Fig.~\ref{figure:appendix:generation_align.}, and finally used as the refined prompt for generation.

\begin{center}
\begin{namedbox}[label=a.3]{User Prompt for Style Alignment.}
Analyze the provided single \{style\_name\} image.

Follow these steps:

1. Identify the artistic medium (e.g., 'oil painting', 'digital illustration').
        
2. Extract 3+ pure visual style elements (focus on color, lighting, texture, mood — avoid any content).
        
3. Describe them in the required format, excluding all subject-related and non-style details.

The output must be STRICTLY $\leq$\{style\_token\_limit\} tokens.
        
Prioritize retaining 3+ key style elements over redundant words if approaching the limit.
        
Use concrete, art-specific vocabulary (e.g., 'impasto texture', 'muted complementary palette', 'film-grain aesthetic').

Output ONLY the following line (no extra text):

 "[artistic medium] with [style element1], [style element2], [style element3+]"
\end{namedbox}
\end{center}

\subsection{Details of Evaluation}\label{appendix:details_of_eval}
In this section, we describe the details of the evaluation metrics used in our experiments.
\begin{center}
\begin{namedbox}[label=a.3]{Prompt for Gemini Feedback.}
YOU ARE A STYLE TRANSFER EVALUATION EXPERT. YOUR SOLE OUTPUT MUST BE A SINGLE JSON OBJECT. DO NOT INCLUDE ANY INTRODUCTORY OR EXPLANATORY TEXT. 

TASK: Assess the images provided in this API call and determine which Candidate Image achieves the most successful and balanced fusion of the "Content Reference Image's" subject structure and the "Style Reference Image's" artistic characteristics. 

INPUT IMAGES: 
\begin{itemize}[noitemsep, topsep=0pt]
\item Content Reference Image (Defines subject, outline, and composition).
\item Style Reference Image (Defines color, lighting, texture, and style).
\item Candidate Images (The images to be compared and scored). 

\end{itemize}

EVALUATION CRITERIA: 
\begin{itemize}[noitemsep, topsep=0pt]
\item Content Fidelity: The accuracy of the candidate image in preserving the subject's identity and shape from the Content Reference (Score 1-5). 

\item Style Fidelity: The accuracy of the candidate image in adopting the color scheme, texture, and artistic attributes from the Style Reference (Score 1-5). OUTPUT REQUIREMENTS: Provide a single JSON object with the following structure. The 'id' field must clearly reference the Candidate Image (e.g., Candidate\_A, Candidate\_1, or the filename if available).
\end{itemize}

\{  "id": "Candidate\_1\_ID", 

"content\_fidelity\_score": 0,      

"style\_fidelity\_score": 0\}

\end{namedbox}
\end{center}

\noindent\textbf{CLIP Score.} Following the procedure in K-LoRA, this metric measures how well each method preserves style. We use a pretrained ViT-B/32 model to encode the generated image and the reference style image into normalized high-dimensional feature vectors. The cosine similarity between the two vectors is then computed to quantify their semantic closeness, which reflects the degree of style preservation.

\noindent\textbf{DINO Score.} This metric is computed in a manner similar to the CLIP Score and evaluates content preservation. The difference is that the encoder is replaced with DINOv2-Base while all other computations remain the same.
\begin{figure}  
\centering  
\includegraphics[width=0.49 \textwidth]{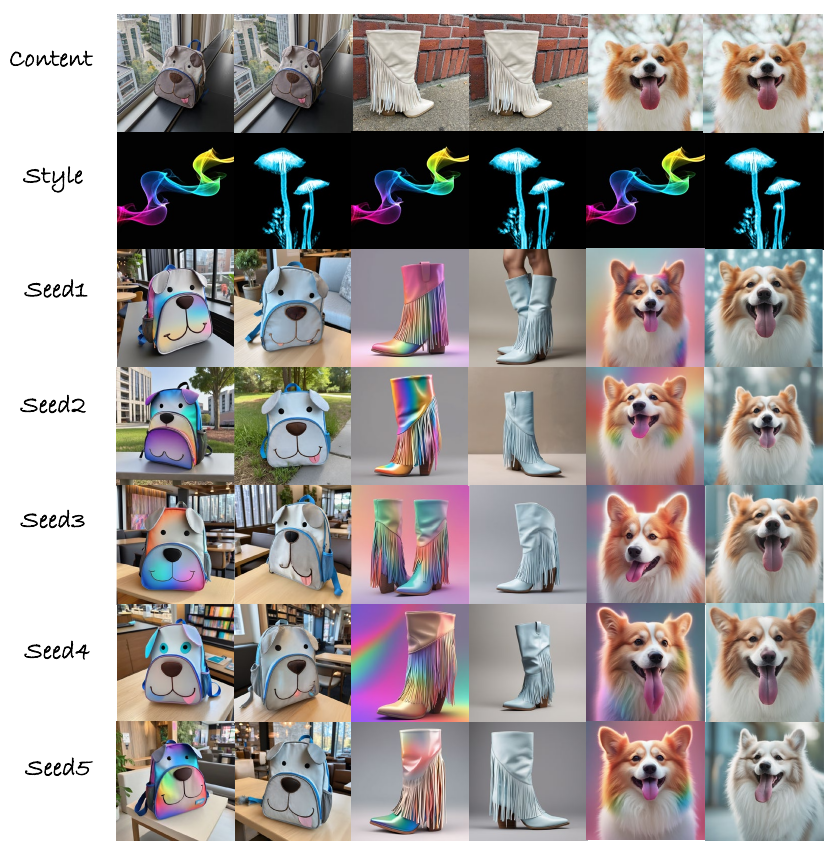}
\vspace{-0.6cm}
\caption{Failure case analysis using SDXL v1.0.}
\label{fig:failure}
\vspace{-0.4cm}
\end{figure}

\noindent\textbf{Gemini Feedback.} For this metric, we query Gemini 2.5-Flash to determine which image among those generated by different methods best aligns with the specified combination of content and style. Gemini assigns a relevance score to each image, and the image with the highest score is selected. Over a batch of results, we compute the selection probability, where a higher probability indicates better generation quality. The reference image and the target prompt are provided as inputs to the model.

\noindent\textbf{Speed.} To evaluate efficiency, we measure the time required to generate ten images for a given content and style pair on a single NVIDIA RTX 3090. For methods that involve training, this time includes both the training phase and the inference phase. For training-free methods, it primarily reflects inference latency. For our method, the total time consists of determining the importance of each LoRA across denoising steps during inference, VLM inference and generating the ten images of each pair.

\section{Discussion}\label{sec:future}
\subsection{Failure Cases}
In this section, we analyze several failure cases of our method. The goal is to identify its current limitations and provide insights for future methodological improvements. Representative failure examples are shown in Fig.~\ref{fig:failure}. For example, the two styles shown in Fig.~\ref{fig:failure}, namely “in abstract rainbow colored flowing smoke wave design” and “in glowing style,” are highly abstract and differ significantly from typical content images. As a result, both styles exhibit varying degrees of style degradation during the training-free optimization process.  We observe that, because our approach does not involve any training, it struggles with highly abstract styles. Generating a coherent combination of such styles and the target content requires a deep understanding of both the prompt semantics and the underlying generative trajectory. However, existing open-source weights may not possess this level of capability, which makes it difficult for training-free methods to achieve satisfactory results in these cases. This suggests that future work may benefit from incorporating lightweight, low-cost training mechanisms to address these limitations.

\subsection{Future Works}

Our analysis of failure cases highlights several promising research directions. Although our framework remains entirely training-free, certain abstract or concept-heavy styles appear to require a deeper semantic understanding than what current open-source weights can provide. Future work may explore lightweight or low-cost training strategies that enhance cross-style and cross-content reasoning without sacrificing the efficiency advantages of our approach. Another interesting direction is to design adaptive refinement modules that dynamically learn style–content interactions during inference, enabling more robust generation in scenarios with highly abstract or visually complex styles. Finally, integrating stronger or more specialized vision-language priors may further improve semantic alignment and mitigate the limitations.

\end{document}